\documentclass[prl,floatfix,twocolumn,showpacs]{revtex4}
\usepackage{amsmath}
\usepackage{amsfonts}
\usepackage{mathrsfs}
\usepackage{epsfig}
\usepackage{graphicx}
\usepackage{dcolumn}
\usepackage{amsmath}

\def\be{\begin{equation}}
\def\ee{\end{equation}}

\begin{document}

\title{Network analysis of a corpus of undeciphered Indus civilization
inscriptions indicates syntactic organization}


\author{Sitabhra Sinha$^1$, Ashraf Md Izhar$^1$, Raj Kumar Pan$^{1,2}$ and 
Bryan Kenneth Wells$^1$}

{\affiliation{%
$^1$ The Institute of Mathematical Sciences, CIT Campus, Taramani,
Chennai 600113, India\\
$^2$ LCE, Helsinki University of
Technology, P.O. Box 9203, FIN-02015 HUT, Finland
}
%\texttt{sitabhra@imsc.res.in}}

\begin{abstract}
Archaeological excavations in the sites of the Indus Valley
civilization ($2500-1900$ BCE) in Pakistan and northwestern India have
unearthed a large number of artifacts with inscriptions made up of
hundreds of distinct signs. To date, there is no generally accepted
decipherment of these sign sequences, and there have been suggestions
that the signs could be non-linguistic. Here we apply complex network
analysis techniques to a database of available Indus inscriptions,
with the aim of detecting patterns indicative of syntactic
organization.
Our results show the presence of patterns, e.g., recursive structures
in the segmentation trees of the sequences, that suggest the
existence of a grammar underlying these inscriptions.
\end{abstract}

%\begin{keyword}
%Linguistic corpus \sep Indus civilization \sep Network analysis
%

\maketitle

%%%%
\newpage
%%%%

% \linenumbers

\section{Introduction}
The last decade has seen a rising interest in the analysis and modeling
of complex networks occurring in many different contexts~\citep{Newman03}, which includes
networks defined in corpora of textual units \citep{Mehler08}. 
Using the graph-theoretic paradigm to study a complex system has often
revealed hitherto
unsuspected patterns in it. While graph-based representation of texts
has
been used for some time in natural language processing tasks, such as,
text parsing, disambiguation and clustering~\citep{Radev08}, the
approach based on the new science of complex networks often asks
questions from a different perspective that can shed new light on the
organization of linguistic structure. For example, networks
constructed on the basis of co-occurrence of words in sentences have
been seen to exhibit (a) the small-world effect, i.e., a small average
distance between any pair of arbitrarily chosen words, and (b) a
scale-free distribution of the number of words a given word is
connected to (i.e., its degree)~\citep{Cancho01}. These properties
have been proposed as reflecting the evolutionary history of lexicons
as well as
the origin of their flexibility and combinatorial nature. A more
recent study of a lexical network of words which are phonological
neighbors has found that the degree distribution might be better fit
by an exponential rather than a power-law function~\citep{Vitevitch08}.
A theoretical model for such word co-occurrence network, which treats
language as a self-organizing network of interacting words, has led to
the suggestion that languages may have a core (the ``kernel lexicon'')
that does not vary as the language evolves~\citep{Dorogovtsev01}.

However, even though text and speech are sequential, focusing exclusively on 
the local correlation between immediately consecutive words may not be a good strategy
to describe natural languages. This is because of the presence of non-local relations
between words that occur apart from each other in a sentence. 
Therefore, network analysis
has been extended to syntactic dependency networks, where two words
are connected if they have been related syntactically in a number of
sentences~\citep{Cancho03}. The theory of complex
networks has also been used to investigate the structure of meaningful
concepts in the written texts of
individual authors, which have been seen to have small-world, as well
as, scale-free characteristics \citep{Caldeira06}. The conceptual
network of a language has been explored by using the semantic
relatedness of words as defined by a thesaurus, and this network too
is seen to have small-world nature with scale-free degree 
distribution~\citep{Motter02}. 

Almost all the network studies done on corpora of textual units so far
have been confined to languages that are still in use.
However, we have historical evidence of many extinct languages, the
knowledge about which have come down to us in the form of
written inscriptions. It is important to consider applying network 
analysis techniques
to such texts and see whether it reveals new insights on the language
as well as the writing system used for it. This is especially so, as
the relation between a language and its writing system is neither
simple nor unique~\citep{Coulmas03}.
While, on one hand, the same language can be written using multiple
writing systems,
on the other hand the same writing system can be used for writing
many different languages. While most network studies have focused on
alphabetic writing, there are
many writing systems (including many that were used recording
languages that are now extinct)
that are based on other principles. These systems may differ remarkably in their
ability to record the various aspects of speech: for example,
logographic writing omits the phonemic structure of speech, while,
phonographic writing may omit vowels and fail to distinguish various
classes of consonants~\citep{Trigger04}. It is therefore intriguing to
consider whether network analysis can reveal the similarities and
differences between such distinct systems of writing, and moreover, if
it can be used to distinguish structural features characterising
writing (i.e., any system of recording language by visible or tactile
marks~\citep{Coulmas03}) from non-writing. 

This is especially
important, as it is clear from observing many of the earliest
examples of writing that have been deciphered, that ``no writing
system was invented, or used early on, to mimic spoken language or
to perform spoken language's function''~\citep{Cooper04}. Instead,
writing was used to record information such livestock or ration
accounts, land grants, offering lists, lexical lists, divinations,
etc., whose storage by verbal or spoken means was difficult and
unreliable; thus the principal function of early writing was
decontextualization and storage~\citep{Goody77}. In almost all cases,
a writing system became more or less capable of expressing spoken
language only after centuries of development, a process of development
that can be clearly seen in Sumerian cuneiform, Egyptian and Mayan
writing~\citep{Cooper04}. It is therefore important to broaden the
results of network analysis of linguistic corpora by applying such
analytical techniques to inscriptions recorded using
different writing systems, and, even to undeciphered inscriptions
for which the underlying writing system is unknown.

In this article, we look at a corpus of inscriptions obtained through
archaeological excavations carried
out in the ruins of the Indus valley civilization. The inscriptions
are in the form of short linear sequences of signs,
of which there are a few hundred different types. Ever since their discovery in the early part of
the 20th century, there have been attempts at deciphering these inscriptions. However, to date there has been no
generally accepted method of interpreting them. We analyze a comprehensive database
of these sequences using techniques inspired by complex network theory. Our aim is to
see whether such methods can reveal the existence of patterns suggesting syntactic
organization in the sign sequences. In the next section, we
briefly introduce the historical context of the Indus inscriptions, while in Section 3, we discuss
the data-set on which analysis has been carried out. Our results are reported in Section 4, and we
finally conclude with a discussion of unresolved questions and further work that needs to be carried
out.

\section{The Indus inscriptions}
\begin{figure}
\centering
\includegraphics[width=0.99\linewidth,clip]{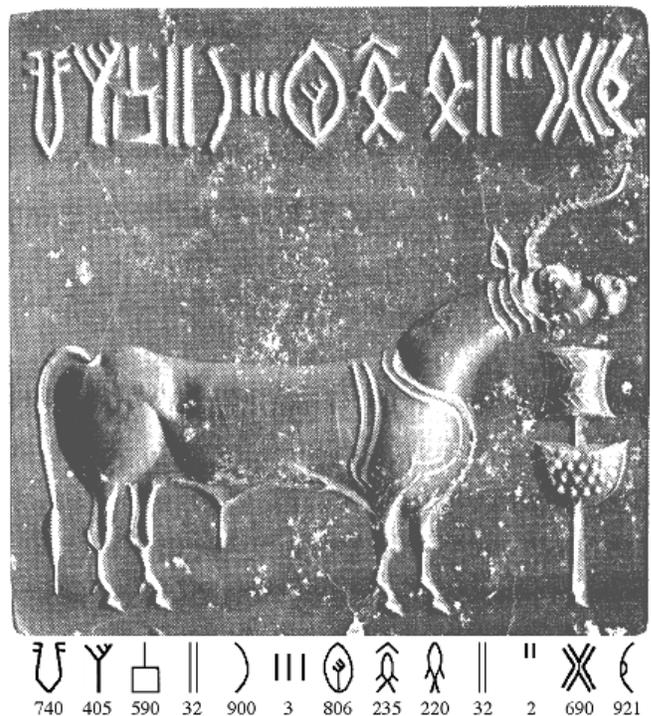}
\caption{Impression made from Seal no. H99-4064 obtained in Harappa
(Mound F, Trench 43) by the HARP team. The inscription at the top is one of the
longest sequences in the WUCS data-set, being 13 signs long. The lower
panel shows the inscription separately, with the ID number for each
sign given according to the W09IMSc sign list. The seal is 
made of steatite,
having dimensions 5.7 cm $\times$ 5.7 cm.
Photograph courtesy of the Harappa Archaeological Research Project, Harappa
Excavations 1999, unpublished preliminary report by
R.~H. Meadow, J.~M. Kenoyer and R.~P. Wright (Figure 32.07).
}
\label{h99-4064}
\label{}       % Give a unique label
%\vspace{-0.15cm}
\end{figure}
The Indus civilization, also known as the Mature Harappan civilization
($2500-1900$ BCE), was
geographically spread over what is now Pakistan and northwestern India, covering approximately
a million square kilometers~\citep{Possehl02}. It was marked by urbanization centered around
large planned cities, as seen from the ruins of Harappa and Mohenjo-daro. Craft specialization
and long-distance trade with Mesopotamia and Central Asia have been well-demonstrated. This
civilization came to an end early in the 2nd millennium BC. There were no historical records of
its existence until archaeological excavations in the late 19th and
early 20th centuries uncovered
artifacts and some of the ruined urban centers~\citep{Marshall31}.
Among the artifacts uncovered during these discoveries were a variety
of objects (especially
seals) that were inscribed with a variety of signs arranged in sequences (Fig.~\ref{h99-4064}). Although found
primarily on seals and their impressions (sealings), inscriptions with similar signs have also
been discovered on miniature tablets, pottery, copper tablets, bronze implements, etc. Unsurprisingly,
given the high sophistication of the civilization and the level of social complexity it
implies, with the concomitant requirements of coordination and communication, these inscriptions
have been interpreted as corresponding to writing. However, despite periodic claims about
decipherment of this writing system, there has as yet been no generally accepted interpretation of
the signs. Despite the lack of success in understanding what the sequences represent, it has been
generally agreed that the inscriptions were written from right to left in the vast majority of 
cases, although a few examples written from left to right or top
to bottom or in boustrephedon are
known~\citep{Mahadevan77,Parpola94,Wells10}.

The failure of decipherment is partly due to lack of knowledge about the language
which the signs encode and the lack of any bilingual texts such as the Rosetta stone which was
crucial in deciphering Egyptian hieroglyphs. While there is disagreement on the exact number
of unique and distinct signs that occur in the inscriptions,
there is overall agreement that they lie in the range of a few
hundred. This rules out the
possibility that the signs belong to (a) an alphabetic system, which contains on average
about 25 letters (such as the Roman or Latin alphabet, used for
writing most European languages including English), (b) a syllabic system consisting of
around 100 signs (such as the kana system for writing Japanese) or
(c) an ideographic writing system (e.g., Chinese), comprising
more than 50,000 characters. However, the number of Indus signs is in the same range
as the number of distinct signs used by logo-syllabic writing systems, such as
Sumerian cuneiform for which the total number of independently
occurring signs has been estimated to be around 900~\citep{Coulmas03}. 
The brevity of the Indus inscriptions (the longest single-line
sequence has 13 signs) and the existence of a large number of signs
that occur with very low frequency
have led to some alternative suggestions regarding the meaning of the sign sequences.
These include the possibilities that, e.g., (i) the signs correspond to a label specifying an individual
and his belongings, in the manner of heraldic badges \citep{Fairservis71} and (ii) the signs
are ritual or religious symbols which do not encode speech nor serve as mnemonic devices,
much as the Vinca signs or emblems in Near Eastern artifacts \citep{Farmer04}. The latter
possibility implies the absence of any syntactic structure in the Indus inscriptions, a possibility
that can be tested without making any a priori assumptions about the meaning of the signs.

\section{Data Description}
Our earlier analysis~\citep{Sinha09} had been done on a corpus of Indus
civilization inscriptions that had been compiled in the process of 
constructing the 1977
electronic concordance of Mahadevan~\citep{Mahadevan77} (see also
Ref.~\citep{Yadav08}).
More recently, Wells has compiled a larger and more comprehensive
database (W09IMSc) of all available inscriptions associated with the Indus
civilization~\citep{Wells10}. This was done by exhaustive search of all
available site reports of Indus excavations and the photographic
corpus of Indus seals and inscriptions~\citep{Joshi87,Shah91}. It was
supplemented by Internet searches for unpublished inscriptions and
requests for unpublished material from individual researchers.
Duplicate records of inscriptions from different sources were manually
controlled.
The W09IMSc database consists of sequences recorded from a total of 
3896 artifacts
and identifies 695 distinct signs. Each sign has been assigned a 3-digit
code between 001 and 958. The sign list has been included as an Appendix
to this paper.

To carry out our analysis, we have focused only on complete
inscriptions, i.e., we have excluded all inscriptions which are only
partially readable because of defaced or ambiguous signs or
damage to the artifact. This results in a reduced set of 2393
artifacts. From this set, we only consider those sequences which
occur on a single line. This is done in order to remove the ambiguity
concerning the
interpretation of sequences occurring as multiple lines of signs,
namely, whether the different lines should be considered as
independent sequences or whether it is one continuous sequence.
Note that we do include artifacts which have sequences (in a single
line) written on multiple sides. In this case, the sequence on each
side is considered separately. Finally, each distinct sign sequence is
considered only once. This series of operations gives us the Wells
Unique Complete Single line text (WUCS) data-set, consisting of 1821
sequences comprising 593 unique signs. All sequences are standardized
to read from right to left. 

\begin{figure}
\centering
\includegraphics[width=0.99\linewidth,clip]{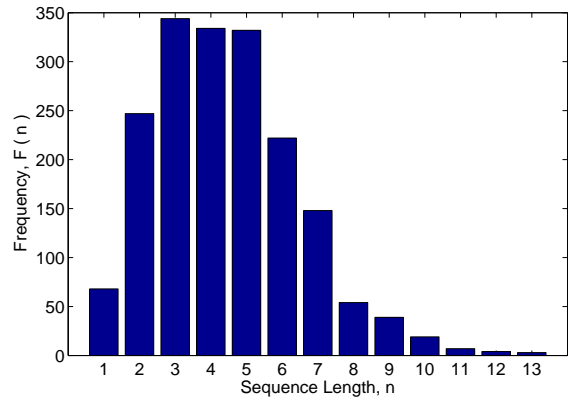}
\caption{Frequency distribution of sequences having different lengths,
$n$, in the WUCS data-set. 
 }
\label{sequence_length_distrn}
%\vspace{-0.15cm}
\end{figure}
The sequences vary in length from 1 to 13 signs, the median length being
4 signs.
Fig.~\ref{sequence_length_distrn} shows the distribution of the
sequence lengths. Many of the single-sign inscriptions (i.e.,
sequences of length 1), are either a solo sign as defined in
the next section, or, appear on one side of an artifact having inscriptions on multiple
sides. A few of the single-sign inscriptions are graphically complex signs that appear to be 
ligatures of two or more relatively simpler signs.

\section{Results}
\subsection{Directed network of Indus signs}
Using the WUCS data-set, a directed and weighted network of the 593 signs can be constructed, where
a directed link from node $i$ to node $j$ implies that sign $j$ occurs
immediately to the left of sign $i$
in at least one sequence. The link from $i$ to $j$ is weighted by the frequency
of occurrence of the ordered pair in the entire data-set.
A distinguishing property of a network having directed connections is the
reciprocity of the links. This can be measured as the ratio of
the number of bidirectional links $L_{bi}$ relative to the total number of
links $L$~\citep{Garlaschelli04}, 
\begin{equation}
R = L_{bi} / L.
\end{equation}
This measure
represents the average probability that if a directed connection exists from
one node to another, the connection in the reverse direction also
exists. For a network where all the links are strictly one-way, $R = 0$,
while, if there is complete absence of directionality, $R = 1$.
In the context of a linguistic network, low values of $R$ would indicate the presence
of significant directional relations between signs, i.e., certain signs are more likely
to appear before (after) certain other signs than after (before) them. Thus $R=0$ would
imply an extremely rigid relation between the signs: if a sign is seen
to appear before
another sign in one context, it will never appear after the other sign in any other context.
For the empirical data-set we calculate $R = 0.148$. 

To measure the significance of the properties calculated for the empirical
data-set we compare them against an ensemble of randomized data-sets. These are generated
by randomly permuting the sign order in each of the 1821 sequences of
the WUCS data-set individually
and re-constructing the corresponding network of sign relations. This is done many times in
order to generate a randomized ensemble of networks. 
Note that the sign frequency of each
sign in a randomized corpus is unchanged from that in the
empirical data-set by construction. Thus,
properties which have almost identical value for the empirical data-set and the randomized ensemble
can potentially be explained as the result of the sign frequency distribution and the fact that certain
signs do not occur together in the same inscription, rather than being a result of the existence of 
syntactic structure, i.e., a set of rules which govern how the signs are consecutively strung
together to form a sequence.
For example, for the 
randomized data-set we obtain the reciprocity, $R_{rand} = 0.338 \pm 0.007$ 
(averaged over 100 realizations). 
As expected, the reciprocity for the randomized networks is much
higher than that for the empirical network, as the shuffling
of sign order in each seal disrupts any existing directional relations
between the
signs in a sequence. 

An alternative randomization is also possible where all the different
sequences are considered together, i.e., all the
signs belonging to every sequence are mixed together and then
reassembled into random sequences. However, this will
generate many random sequences with signs that never co-occur in
the same sequence in the empirical data (i.e., in the
inscriptions). Thus, the first randomization method, by taking
into account the context in which two signs co-occur,
gives a much stricter criterion for deciding which of the features
of the empirical network are significant (i.e., unlikely
to appear by random chance given the frequency of occurrence of
each sign).

\subsection{Comparison of empirical properties with randomized ensemble}
A preliminary analysis of the data shows that 21 signs only appear as
{\em solo} (or single-character) inscriptions and never in conjunction with
another sign, viz., sign numbers 037, 039, 047, 110, 147, 281, 341, 386, 387, 699,
753, 780, 781, 782, 823, 841, 942, 945, 946, 956 and 957. 
In network terminology, the nodes corresponding to these signs have no
in-coming
links (i.e., in-degree$=0$) or out-going links (i.e., out-degree$=0$). 
Also, out of the 369 distinct signs that can appear at the start of a sequence of length $\geq 2$, 128 signs are only seen at the beginning and never in any other position in a sequence.
In network terminology, these signs have no in-coming links (i.e., in-degree=0).
We shall call these signs ``beginners": 
025, 027, 028, 029, 041, 046, 051, 057, 058, 059, 069, 084, 098, 107, 112, 114, 117, 118, 119, 121, 126, 131, 133, 141, 144, 145,
146, 166, 178, 195, 201, 208, 209, 216, 229, 230, 261, 262, 266, 272, 276, 319, 323, 
325, 327, 329, 361, 363, 370, 389, 403, 412, 428, 445, 451, 458, 473, 479, 490, 493,
501, 505, 513, 528, 541, 544, 545, 551, 563, 571, 573, 577, 579, 586, 591, 601, 620,
622, 625, 631, 634, 635, 640, 641, 678, 681, 683, 687, 688, 689, 693, 694, 698, 707,
710, 716, 728, 731, 736, 747, 751, 764, 768, 777, 795, 796, 799, 815, 818, 826, 827,
829, 843, 852, 859, 863, 864, 870, 871, 876, 878, 891, 896, 902, 903, 918,922 and 950. 
Similarly, out of the 196 distinct signs that are seen to terminate a sequence in the
WUCS database, 43 signs are seen only at the end of a sequence and never in any other position. 
In network terminology, these signs have no out-going links (i.e., out-degree=0)
These signs
will be referred to as ``enders": 045, 074, 105, 106, 129, 138, 157, 173, 200, 203,
224, 256, 289, 294, 303, 307, 324, 375, 394, 409, 410, 423, 427, 429, 430, 481, 512, 712, 749,
822, 842, 855, 860, 866, 869, 872, 875, 907, 909, 911, 930, 932 and 951.
Note that among the remaining 401 signs (i.e., signs which do not
belong to any of the above three classes) there are many that can
occur at the beginning of a sequence, as well as elsewhere. Similarly
several signs that can be observed to terminate some sequences may
also be seen at other positions in a sequence. Indeed, there are 120
signs that are seen to begin some sequences or end other sequences.
There are thus 127 signs which are seen to be always preceded as well as followed by other signs in any inscription that they occur in.

\begin{figure*}
\centering
\includegraphics[width=0.85\linewidth]{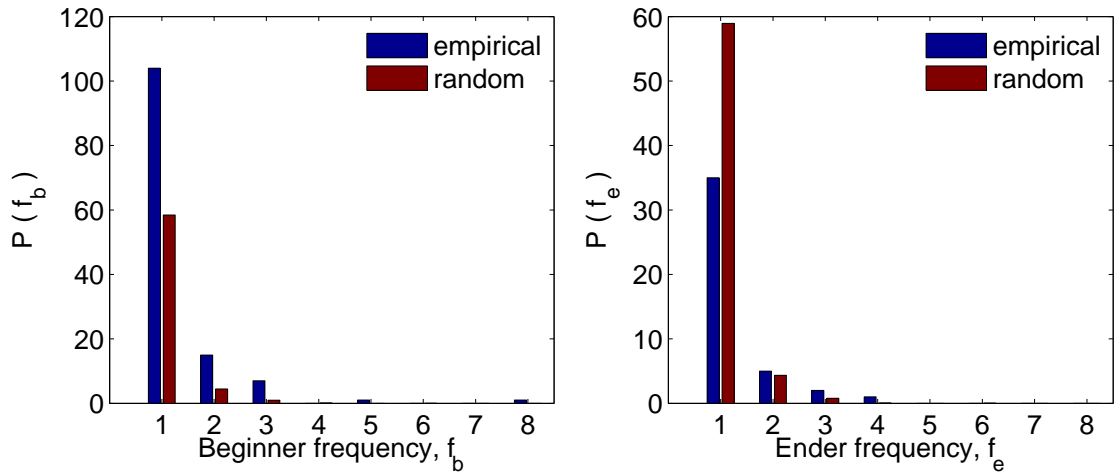}
\caption{The frequency distributions for (left) beginners, $f_b$,
and (right) enders, $f_e$, are shown for the WUCS data-set. The
corresponding distributions for
randomized sequences, averaged over 100 different realizations, are
shown for comparison.}
\label{begin_end_distrn}
%\vspace{-0.15cm}
\end{figure*}
Not surprisingly, the number of signs that appear as solo inscriptions
in the randomized data-sets is the same as
in the original data (= 21). This is because we randomize each sequence separately, and do not put together
signs in a randomly generated sequence if they have never co-occurred in
any of the empirical sequences.
However, the number of beginners and enders will be different,
as these depend on the underlying network relations which have been changed by randomizing
the corpus. For example, calculating the number of beginners and
enders in 100
randomized data-sets yields $n_b = 66.26 \pm 6.59$ and $n_e = 62.61
\pm 6.71$, respectively.
It implies that, in the empirical data, there are significantly more beginners on one hand,
and a significantly lower number of enders on the other, than would be expected purely on the basis
of chance, given the frequency of occurrence of the individual signs.
It is also possible to observe with what frequency beginners or enders
in the empirical data-set
appear in the same role in the randomized corpus. We observe that
four beginners in the empirical set (signs 201, 216, 272 and 545) and two enders (signs 409 and 423)
never occur as beginners and enders (respectively) in the 100 randomized trials we carried out. It implies
that the occurrence of these signs always at the beginning or end of a sequence (and
never in any other position) may be highly significant,
and certainly not a result of simple chance.

We can now ask: in how many different sequences does a particular
beginner or ender appear? In answer, we see that
most beginners or enders occur only in very few distinct sequences.
In Fig.~\ref{begin_end_distrn}, we have shown the frequency
distribution of the beginners (left) and enders (right) in the
empirical data and compared it with that observed from
averaging over the distributions corresponding to 100 randomized trials. 
The difference between the empirical and random distributions is
significant at low frequencies as it is much larger than the standard
deviations for the randomized data.
Note that there are many more beginner signs in each frequency class
than would be expected had all the sequences
in the corpus been randomly scrambled. The largest number of
distinct sequences that a beginner can appear in is 8.
Indeed, most of the beginner signs appear in only $1-3$ distinct
sequences. This is even more so the case for enders,
where a particular ender sign can appear at the end of a maximum
of 4 distinct inscriptions.

\subsection{Degree and strength distribution}
\begin{figure}
\centering
\includegraphics[width=0.99\linewidth,clip]{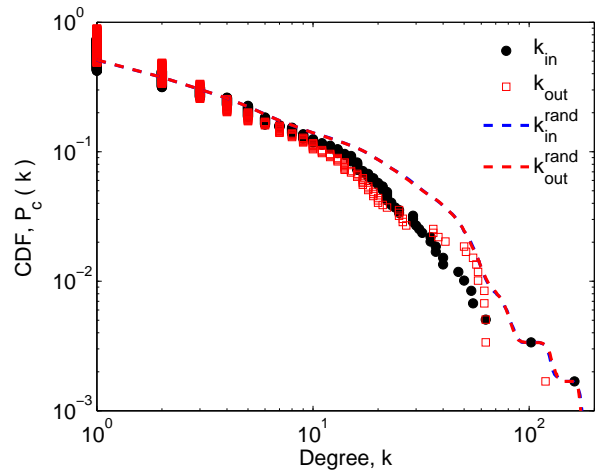}
\caption{The cumulative distribution function for in-degree, $k_{in}$, and out-degree, $k_{out}$, of 
the directed network of signs constructed from the WUCS data-set. 
The corresponding distributions for
randomized sequences, averaged over 100 different realizations, 
are shown for comparison. The standard deviations of the randomized
data over the range $20\leq k \leq 50$ are less than $4 \times 10^-3$.}
\label{degree_distrn}
%\vspace{-0.15cm}
\end{figure}
We now focus on the distribution of the number of links for each sign (i.e., the degree).
If there had been a rigid relation between the signs, i.e., the occurrence of one particular sign was always preceded or
followed by another particular sign, this would show up in the degree distribution. Thus, the occurrence of a
sharply decaying degree distribution with an overall low number of links per sign would indicate that for most
signs there is not much freedom of choice in deciding which sign will precede or follow it.
Fig.~\ref{degree_distrn} shows that both the in-degree (the number of incoming connections) and the
out-degree (the number of outgoing connections) have long-tailed distributions, indicating that there is
relatively a high degree of variation in the signs that a particular sign
occurs adjacent to. However, this also does not
correspond to the total freedom in choosing neighbors as is the case for a 
random sequence.
The  curves for the in-degree and out-degree distribution for the networks constructed from
the randomized ensemble obtained by shuffling sign order in each of
the sequences show a consistently
higher probability for larger degrees. This indicates that the variation of sign relations is much more
restricted in the empirical sequences than would be the case had each
of the sequences been put together
randomly.

\begin{figure}[tbp]
\centering
\includegraphics[width=0.99\linewidth,clip]{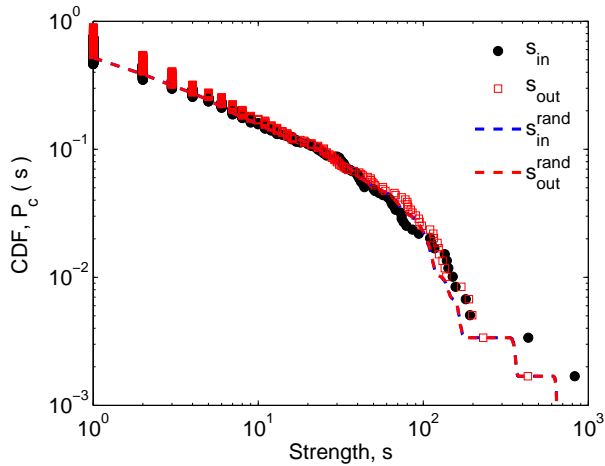}
\caption{The cumulative distribution function for in-strength, $s_{in}$, and out-strength, $s_{out}$, of 
the directed network of signs constructed from the WUCS data-set. The corresponding distributions for
randomized sequences, averaged over 100 different realizations, are shown for comparison. }
\label{strength_distrn}
%\vspace{-0.15cm}
\end{figure}
Another related property that is often observed in the case of weighted networks is the distribution of
strength, i.e., the weighted sum of links for a node. To a certain extent, this is governed by the frequency
of occurrence of individual signs, as a more common sign will have many more relations with other
signs. Not surprisingly, in Fig.~\ref{strength_distrn},
we see that the in-strength and out-strength distributions of the networks
constructed from empirical and randomized data match fairly well. In other words, the strength distribution
of the WUCS network can be explained almost fully on the basis of individual sign frequencies
and the fact that certain signs do not occur together in the same sequence.

%The relative balance between incoming and outgoing connections is
%measured by the fan-in fan-out complexity, normalized by the total degree,
%\begin{equation}
%C_{fan} = \frac{2 \sqrt{k_{in} k_{out}}}{k_{in}+k_{out}},
%\end{equation}
%where $k{in}$ and $k_{out}$ are the in-degree and out-degree
%respectively. A sign that occurs exclusively at the beginning or the
%end will have $C_{fan} = 0$, while if it has the same in-degree as
%out-degree, $C_{fan}=1$.

\subsection{Core-periphery organization}
The connectivity, or average density of connections, in a network is
measured as the ratio of non-zero entries in the
adjacency matrix to the total number of matrix elements.
The network of Indus sign relations is extremely sparse, with a connectivity
of only 0.0077. This may be compared with the connectivity for the
corresponding randomized corpus, $C_{rand} = 0.011$ (averaged over an ensemble
of 100 different realizations, the standard deviation being less than
$10^{-4}$). 
Thus, the WUCS data-set shows that out of a possible
$593 \times 593 = 351,649$ directed
sign pairs, only 2719 sign pairs are actually observed (as compared to
the $3827 \pm 20$ directed signs pairs observed when the corpus is
randomized). This already suggests the existence of grammatical rules
in the construction of the sequences that prevent the occurrence of a
vast majority of the possible sign pairings.

\begin{figure}
\centering
\includegraphics[width=0.99\linewidth,clip]{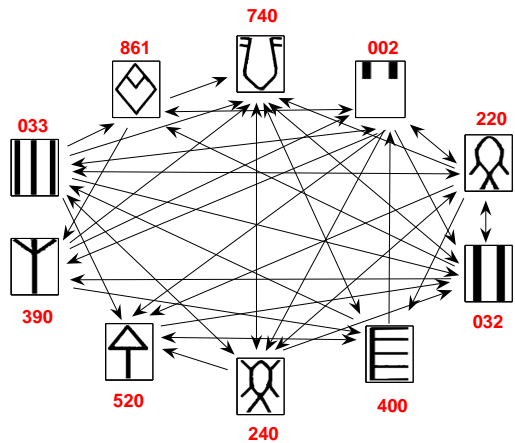}
\caption{The sub-network of the ten most frequently occurring signs in
the WUCS data-set, with the
ID numbers of each sign indicated alongside its image. Note that the connectivity in this subnetwork
is substantially more dense ($\sim 0.49$), with about half of all the potential connections present, relative
to the entire network whose connectivity is 0.0077.
}
\label{ten_most_freq_signs_network}
%\vspace{-0.15cm}
\end{figure}
\begin{figure}
\centering
\includegraphics[width=0.99\linewidth,clip]{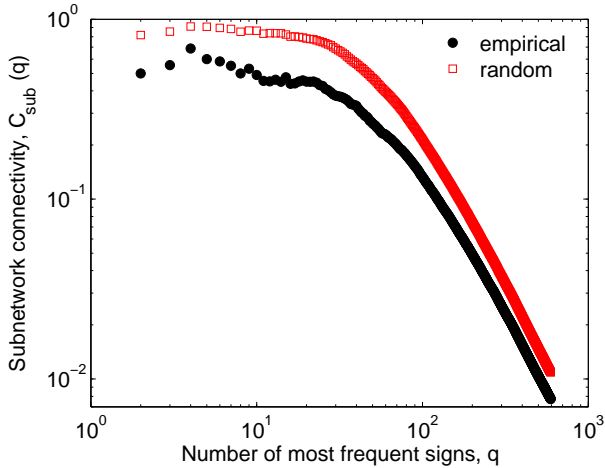}
\caption{The connectivity for the sub-network of $q$ most frequently occurring signs
in the WUCS data-set ($q = 2, 3, \ldots, 593$) for the network constructed from the empirical data, compared to that
for the networks constructed from randomized data-sets. 
The higher connectivity for the randomized case is an outcome of the
long-tailed nature of the distribution of frequencies of individual signs.
The values for the randomized data are averaged over 100 different
realizations. The error bars are not indicated as they are smaller
than the symbol size used for the randomized data.
}
\label{subnetwork_connectivity}
%\vspace{-0.15cm}
\end{figure}
If we graphically represent the sub-network of connections between
nodes corresponding to the 10 most
common signs in WUCS (i.e., the ones occurring with the highest
frequency), we note that
they are strongly inter-connected (Fig.~\ref{ten_most_freq_signs_network}). In fact, almost half of 
all the possible sign pairs in this sub-set are actually observed to occur. This is partly an outcome of the
inhomogeneous frequency distribution of the individual signs, with the most frequently occurring
signs appearing in many different sequences and thereby having
connections with a large variety
of signs. This is indicated by a comparison of the connectivity of the sub-network of $q$ most frequent
signs ($q$ ranging from 2 to 593) for the empirical network with that for the randomized ensemble. As
Fig.~\ref{subnetwork_connectivity} shows, the connection density for
the sub-set of most frequently
occurring signs is much higher than what would have been expected had
the signs been placed in a sequence randomly (based only on
their individual frequency of occurrence and the restriction that
certain signs never co-occur
in the same sequence). The lower sub-network density for the empirical network is a result of several
possible sign relations (which have a very high probability of occurring in a randomized sequence) 
never appearing in the WUCS data-set. It suggests the existence of syntactic relations in
constructing the sequences that prevent the occurrence of these highly probable sign relations.

The above analysis also indicates that there exists a core set of
signs that appear together very frequently in
a sequence.
A natural question is whether the network generated from the WUCS
data-set has a core-periphery
organization. This would imply the existence of a densely connected
central core (central in terms of network distance between the nodes)
and a larger class of sparsely connected peripheral nodes, like that
seen in the
case of geographically embedded transportation networks~\citep{Holme05}. To obtain such a decomposition
of the network we use a pruning algorithm that successively peels away the ``outer layers" of peripheral nodes 
to reveal a subnetwork of a
given core-order. The $k$-core of a network is defined as the subnetwork containing
all nodes that have degree at least equal to $k$. Thus, to obtain it, we have to iteratively remove
all nodes having degree less than $k$. In particular, the 2-core of a
network is obtained by recursively eliminating
all nodes that do not form part of a loop (i.e., a closed path through
a sub-set of the connected
nodes). For a $k$-core, there are at least $k$ paths between any pair
of nodes belonging to it. It is
obvious that for any network, there exists an innermost core of a
maximum order that cannot
exceed the highest degree of the network. 

\begin{figure}
\centering
\includegraphics[width=0.99\linewidth,clip]{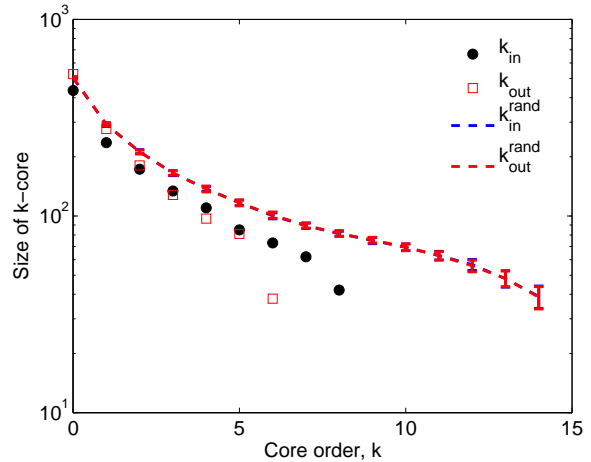}
\caption{The $k$-core decomposition of the directed unweighted network
of Indus signs for in-degree and out-degree,
compared to that for networks constructed from the set of randomized sequences keeping individual sign frequencies
unchanged (averaged over 100 different realizations). Error bars
indicate the standard deviations for the randomized data.}
\label{k_core_size}
%\vspace{-0.15cm}
\end{figure}
In a directed network, one can define a $k$-core
either in terms of the in-degree (number of connections arriving at the node) or the out-degree
(number of connections sent from the node). For the WUCS network, the innermost $k$-core turns
out to have order 6 for out-degree and 8 for in-degree
(Fig.~\ref{k_core_size}).
The corresponding core-size for the networks constructed from
randomized sequences are also shown in Fig.~\ref{k_core_size}. The
values for the randomized data are consistently
higher than those for the empirical network.
The intersection of the innermost out-degree core and the innermost
in-degree core comprises 26 signs that are the ones most likely to occur at the
medial positions of a given inscription: 001, 002, 003, 031, 032, 
033, 140, 220, 231, 233, 235, 240, 368, 415, 590, 700, 705,   
706, 717, 740, 741, 798, 803, 820, 840 and 904. By observing which signs
appear in the innermost out-degree core but not in the intersection
set (i.e., taking the difference of these two sets), we obtain 12
signs that most frequently precede other signs in an inscription:
055, 060, 440, 575, 615, 692, 742, 745, 790, 806, 900 and 920.
Similarly, by considering the signs which appear in the in-degree core
of highest order but not in the intersection set, we obtain 16 signs
that most frequently follow other signs in an inscription: 017, 090,
100, 125, 151, 176, 255, 350, 388, 390, 400, 455, 520,
550, 760 and 861. Together these 54 signs are the ones most
likely to be used in an inscription.

\begin{figure}[tbp]
\centering
\includegraphics[width=0.99\linewidth,clip]{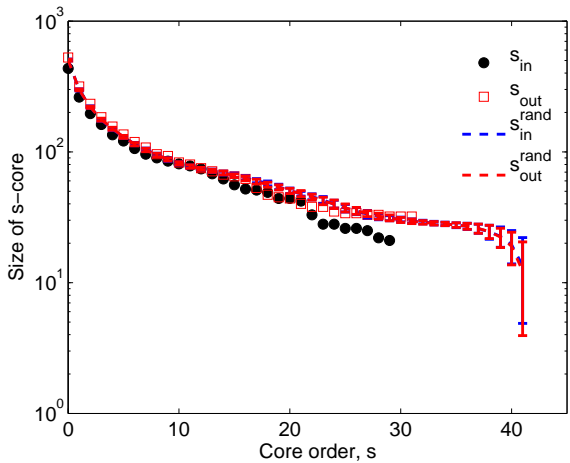}
\caption{The $s$-core decomposition of the directed weighted network of Indus signs for in-strength and out-strength,
compared to that for networks constructed from the set of randomized sequences keeping individual sign frequencies
unchanged (averaged over 100 different realizations). Error bars 
indicate the standard deviations for the randomized data.} 
\label{s_core_size}
%\vspace{-0.15cm}
\end{figure}
We can generalize the concept of $k$-core from the degree to the strength of a node, thereby defining
a $s$-core.
The $s$-core of a network is defined as the subnetwork containing
all nodes that have strength at least equal to $s$. Thus, to obtain
it, we have to iteratively remove
all nodes having strength less than $s$.
Fig.~\ref{s_core_size} shows the core size variation with core-order for both in-strength and out-strength.
The innermost $s$-core for out-strength has order 31, and that for in-strength has order 29.
As is clearly seen, the out-strength core size for the empirical network matches fairly well with that of the randomized  
networks, while the in-strength core for the empirical network is
smaller than that for the randomized
network at large core order, $s$. It implies that the set of mutually
connected signs having high in-strength is significantly smaller
than would be expected on the basis of chance had the signs been placed randomly in each sequence.
This indicates the presence of certain context-based restrictions on
the pairing of signs. In other words, the
occurrence of a sign pair depends on what other signs occur in that
sequence.

As in the case of degree, for strength also we can look at the
intersection of the innermost in-strength and out-strength cores,
which provides us with a set of 18 signs: 001, 002, 003, 032, 033,
100, 176, 220, 233, 235, 240, 390, 415, 590, 705, 740, 798 and 803.
This is a sub-set of the group of signs obtained above by considering
the intersection of the in-degree and out-degree cores of highest
order, excepting for signs 100, 176 and 390. By considering the
difference of the intersection set with the set corresponding to the
innermost out-strength core, we obtain the signs that are
most likely to precede the medial group of signs: 031, 060, 368, 690, 706, 741,
760, 806, 817, 820, 840, 861, 900 and 920. Similarly, from the difference
of the intersection set with the set corresponding to the
innermost in-strength core, the signs that are most likely to follow
the medial group of signs is obtained: 090, 400 and 520. Together
these 35 signs can be considered to constitute the ``core lexicon'' of
the Indus inscriptions.

\subsection{Network of significant links}
\begin{figure}
\centering
\includegraphics[width=0.99\linewidth,clip]{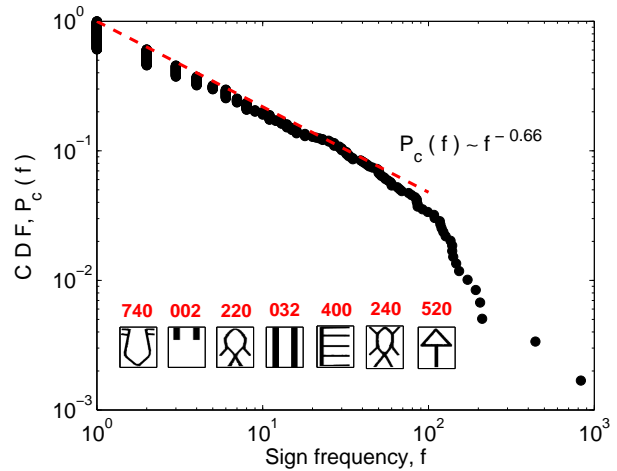}
\caption{The cumulative probability distribution function, $P_c$, of the different sign frequencies, $f$. The broken
line shows a power law fit of the data, $P_c (f ) \sim f^{-\alpha}$, with the maximum likelihood estimate 
for $\alpha$ over the range [1,30] being 0.66. This implies that the corresponding Zipf exponent is 
$1/\alpha \simeq 1.5$. The seven most frequently occurring signs in
the WUCS data-set are shown as insets,
along with the corresponding ID numbers,  
arranged in decreasing order of frequency from left to right.
 }
\label{sign_freq_distrn}
%\vspace{-0.15cm}
\end{figure}
So far we have placed all sign pairs that occur in the WUCS data-set on an equal footing. However, certain
pairs may occur with high probability simply because the individual signs that make up the
pair occur with high frequency. Fig.~\ref{sign_freq_distrn} shows that the frequency distribution of sign occurrences 
in the WUCS data-set has an approximately power-law 
distribution~\citep{note1}
This implies that the most common signs will occur in a very large number of
sequences (the most frequent sign ``740" appearing as many as 831 times, which is more than $10 \%$ of the total of 8095
occurrences of the 593 signs in the WUCS data-set). By using the information about the probability of occurrence
for individual signs in the data-set we can investigate significant sign relations, i.e., sign
combinations that occur far more frequently than would be expected from the individual probabilities of
the component signs.
Thus, if sign $i$ occurs with a probability $p (i)$ and $j$ with $p (j)$ in the corpus, then the pair $ij$ is significant if it
occurs with a probability
$p (ij) \gg p (i) p (j)$. If $p (ij) \simeq p (i) p (j)$, we can
conclude that the two signs are essentially independent
of each other, and their joint occurrence is not indicative of any significant relation between them.
To measure by how much $p (ij)$ has to be larger than the product of $p(i)$ and $p(j)$ in order to
be significant, we need to compare the empirical joint occurrence
probability against the corresponding
value for the randomized ensemble. The randomized corpus is generated
(as described earlier) by shuffling the
order of signs in each sequence, such that the pair correlations in the original data are removed while keeping the
individual sign frequencies unchanged.
A sign pair $ij$ is considered significant if the empirical
relative frequency of its occurrence, $P_{emp} (ij)$, is so large compared to the corresponding relative frequency
in the randomized corpus, $P_{rand} (ij)$, that the pair can never occur with
the observed frequency had the two
signs been independent, i.e., had there been no
dependency relation between them. This deviation of the empirical pair probability from that corresponding
to the randomized corpus
can be quantified by computing the $z$-score:
\begin{equation}
z_{ij}= \frac{P_{emp} (ij) - \langle P_{rand} (ij) \rangle}{\sigma_{rand}(ij)},
\end{equation}
i.e., the difference between the
relative frequencies of the sign pair for the empirical data and the mean for the randomized ensemble,
divided by the standard deviation of the frequencies obtained for the
different randomizations.

\begin{figure}
\centering
\includegraphics[width=0.99\linewidth,clip]{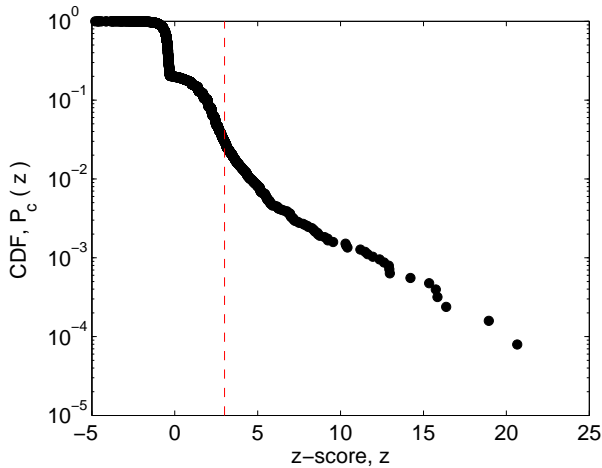}
\caption{The cumulative distribution function of the $z$-scores for
all sign pairs in the WUCS data-set. The vertical
broken line represents a $z$-score of 3, indicating that if the distribution had been Gaussian then all pairs having $z$-score
to the right of this would have been considered ``significant". 377 sign pairs are observed to have $z > 3$.}
\label{z-score_cdf}
%\vspace{-0.15cm}
\end{figure}
The cumulative probability distribution for the $z$-scores of all sign pairs are shown
in Fig.~\ref{z-score_cdf}. Had this distribution been a Gaussian, all sign pairs with $z$-scores higher than 3 could have been
considered significant. The empirical distribution is observed to have
a long tail and we can consider all pairs to be significant 
which have a $z$-score higher than a specified cut-off, $z_c$. We note that there are 377 sign pairs
with $z$-score larger than $z_c = 3$, while the 31 significant pairs
obtained when $z_c = 8$ are shown
as a network of ``most significant relations" in Fig.~\ref{significant_links_zscore_greater_than_8}.
This comprises 36 signs, containing 25 out of the 30 most frequent signs, indicating that some of the 
commonest signs have significant relations between them. While
most significant pair relations are between such common signs, one notable exception is the significant
relation between sign ``017" (46th most common sign) and sign ``585" (67th most common sign), 
both of which are relatively low-frequency signs.
As this sign relation has a very high $z$-score, although the
individual signs are themselves not very common,
it is an intriguing sign pair and possibly has some functional significance in terms of interpreting
the sequences.
\begin{figure}
\centering
\includegraphics[width=0.99\linewidth,clip]{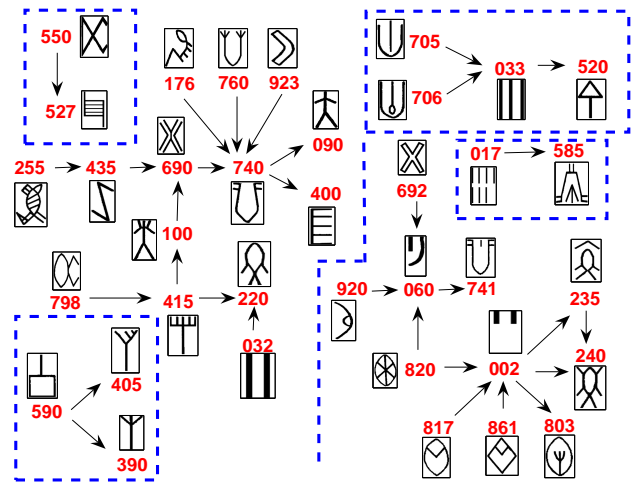}
\caption{The network of 31 significant sign pairs having $z$-value $> 8$, obtained by comparing with an ensemble of
randomized sequences. The $z$-score is calculated
by using the mean and standard deviation of the relative frequencies of each sign pair obtained from an ensemble
of $10^4$ randomized sets of the WUCS data. There are a total of 36 signs grouped into six isolated sign clusters, 
each indicated by a broken boundary.}
\label{significant_links_zscore_greater_than_8}
%\vspace{-0.15cm}
\end{figure}

\subsection{Syntactic tree generation}
\begin{figure}
\centering
\includegraphics[width=0.99\linewidth,clip]{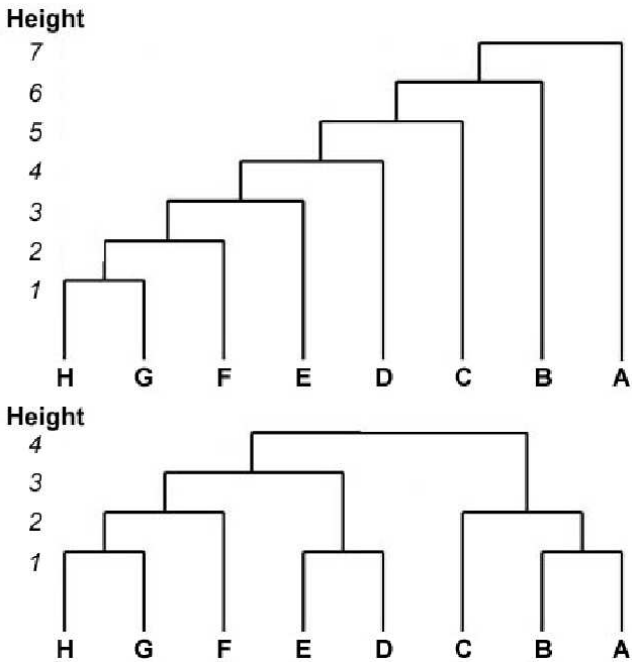}
\caption{Schematic segmentation trees for a sign sequence of length 8, representing two alternative
possibilities. The top example is a relatively unstructured sign sequence, with the tree height being
almost identical to the sequence length. The bottom example shows significant recursive structure
and a corresponding lower tree height.}
\label{trees}
%\vspace{-0.15cm}
\end{figure}
We shall now attempt to reveal recursive structure indicative of the
presence of syntactic rules for generating the inscriptions by
``parsing" the longest sign sequences. We do this by generating
segmentation trees of the sign sequences based on the statistical
significance of sign pair occurrences. Given a inscription of length
$n$, sign pairs are successively merged in decreasing order of their
statistical significance, with the first merger being done for the
sign pair with the highest $z$-score in that sequence.
The next merger is done for the pair of signs having the next highest
$z$-score and so on, until all the signs in the sequence have been
merged. In case of a tie between two or more pairs at any stage, the
leftmost pair is chosen. This sequence of mergers is then ``unfolded''
to produce the resulting segmentation tree of the sign sequence which
is shown schematically in Fig.~\ref{trees}. The height of the tree is
an indicator of the presence of significant recursive structure in the
sign sequence. In particular, if the signs are all independent of each
other, then the segmentation tree has essentially the same height as
the length of the sequence (Fig.~\ref{trees}, top). On the other hand,
if for long sequences there exists subsequences that appear as an
unit in the corpus several times, including as complete
sequences in their own right, this is
indicative of recursion. 
The corresponding tree height is
substantially reduced as compared to the sequence length
(Fig.~\ref{trees}, bottom).

\begin{figure*}
\centering
\includegraphics[width=0.85\linewidth]{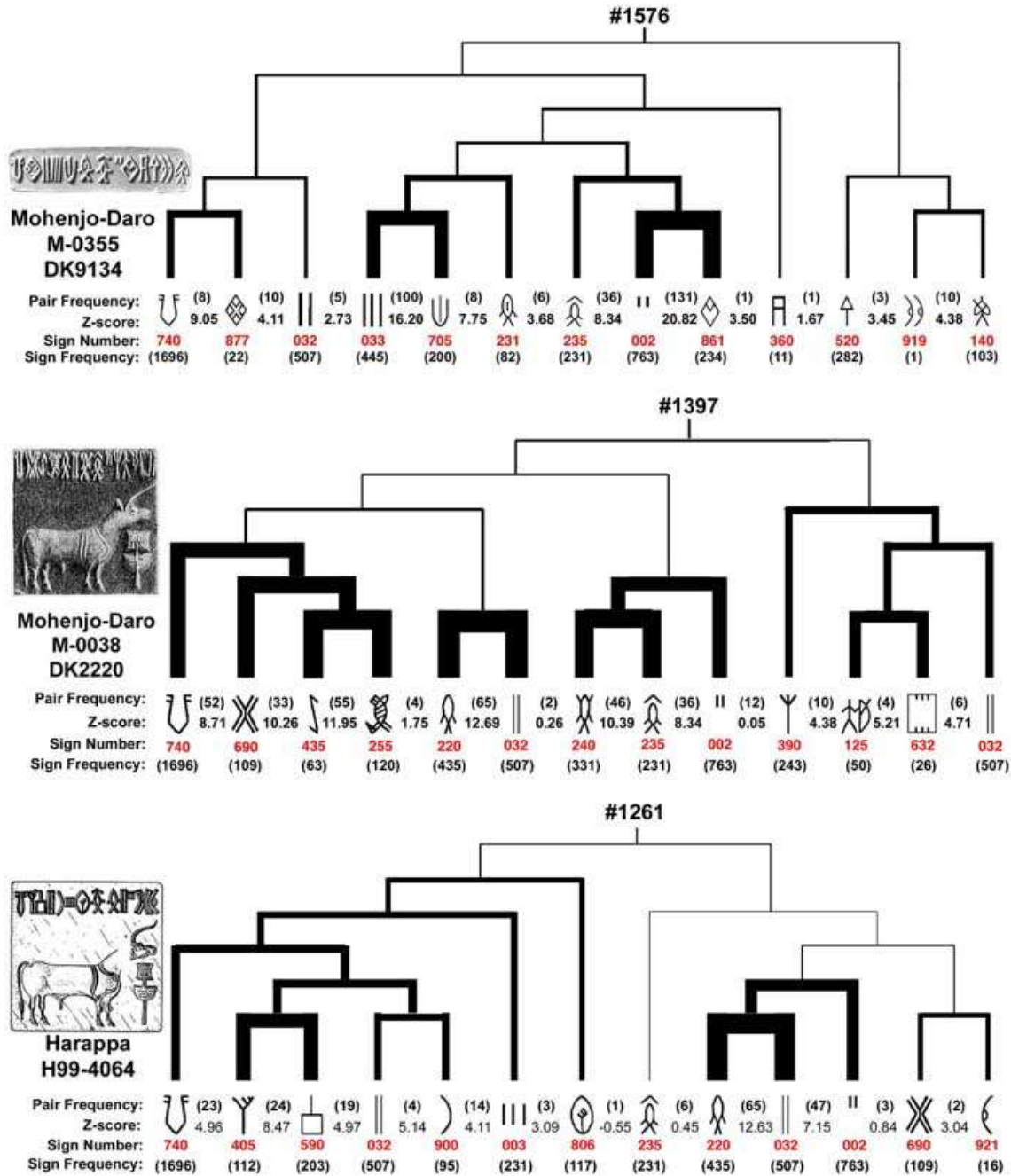}
\caption{The segmentation trees for the three longest sequences in the
WUCS data-set, viz., the inscriptions of Seal M-0355 from
Mohenjo-Daro (top), M-0038 from Mohenjo-Daro (center)
and H99-4064 from Harappa (bottom), obtained by using the
$z$-scores of the 12 sign pairs comprising each of these sequences as
described in the text. The thickness of the lines are proportional
to the corresponding $z$-score values.
At the left of each tree, an image of the corresponding seal is shown.
The $z$-score computed for each pair, as well as the 
corresponding pair 
frequency in the W09IMSc data-set, are indicated below each pair.
Also shown is the total frequency of occurrence of each constituent
sign in the
W09IMSc data-set.}
\label{segmentation_tree}
\end{figure*}
We use this criterion to seek a signature of recursive, and hence
syntactic, structure in the WUCS data-set.
We have confined our attention to parsing the 33 inscriptions in WUCS
data-set having 10 or more signs. Our earlier analysis of
the EBUDS data-set~\cite{SInha09} had shown that the average
tree height for the longer sequences was around
5.
We had concluded that the existence
of such a characteristic length scale indicated that the longer sequences were actually composed of multiple
smaller sequences that can occur independently in the corpus, and which have definite syntactic
relations among their constituent signs. This is confirmed by the
recent analysis done on the WUCS data-set.

Fig.~\ref{segmentation_tree} shows the segmentation trees
of the three longest sequences, each  comprising 13 signs. Two of
these, M-0355 (sequence no. 1576 in the WUCS database) and M-0038
(sequence no. 1397 in the WUCS database) clearly indicate that they are
made up of 3, or possibly 4, sub-sequences,
while the third, H99-4064 (sequence no. 1261 in the WUCS database)
appears to comprise two long sub-sequences.
Let us consider the first of the sequences, M-0355.
The 3-sign cluster ``520-919-140" at the beginning of the sequence is the initial phrase, and is separated from
the rest by sign 360. The medial sequence is broken into two parts
``235-002-861" and ``033-705-231".
The sequence ends with the terminal phrase ``740-877-032". We observe that each of these four
sub-sequences obtained by this analysis also occur as units in other
inscriptions in the WUCS data-set,
thereby verifying the accuracy of the segmentation procedure. By breaking down long texts into (possibly meaningful)
phrases that have independent existence, the method should help in identifying the grammatical rules by which
the sequences are written.

\section{Discussion}
In this paper we have used complex network analysis techniques on the sign network constructed
from a sub-set of the corpus of inscriptions obtained in Indus civilization excavations.
Our results suggest that though these sign sequences are yet to be deciphered, they have a
highly structured arrangement which is suggestive of the existence of syntax. The inference of a
set of rules (i.e., the grammar) for arranging these signs in a particular order, so as to be able
to create pseudotexts that are indistinguishable from the excavated ones, is the eventual aim of
the analysis described here. However, before we can successfully compile the ``grammar" for these sequences, 
several open problems need to be addressed.
One of the extensions of the present work has to do with looking beyond sign pairs to sign triplets,
quadruplets, etc. Preliminary analysis of networks of ``meta-signs" by
us indicates that
combinations beyond four signs may not have statistical significance. A detailed comparison
between the sign network described here and the meta-sign network may provide clues about the
possible hierarchical arrangement of subsequences in the longer sequences. Evidence of this
is already seen from the construction of segmentation trees of individual sequences based on
relative pair frequencies. It is also possible that there are non-local correlations
between signs in a given inscription. To analyze this, we need to redefine the links in the
network as being connections between all signs that occur in the same inscription. Again, preliminary
analysis seems to suggest that this does not provide substantially new results from those
reported here. Based on the number of distinct signs (around $500-600$) there have been several suggestions that, as
the number is too high to be an alphabetic system but too small for
an ideographic system, the
inscriptions could well be written in a logo-syllabic system. Such a writing system combines
both logograms (morphemic signs) and syllabic (or phonetic) signs without inherent meaning. In
future work, we plan to investigate the differences that arise in the network structure of languages
belonging to these very different systems, in order to make an inference on the nature of the
writing system used in the Indus inscriptions. One of the most controversial aspects of Indus
decipherment is the question of how many distinct signs are there. Mahadevan \citep{Mahadevan77} 
identified 417 signs, while Wells \citep{Wells10} has distinguished about 700 signs. Other researchers 
have come up with a wide range of different numbers. Therefore, an important open issue that 
needs to be settled is the robustness of these results with respect to the sign list being used.

However, despite these limitations, based on the results reported here
it seems fair to conclude that the
inscriptions do have an underlying syntactic organization. By comparing with a randomized ensemble
of sequences that maintain the original sign frequency and restrictions on the co-occurrence of signs
in the same inscriptions, but which otherwise lack any of the empirical correlations between sign pairs,
we have established beyond reasonable doubt that the sequences cannot be just random juxtaposition
of signs. It appears to rule out the possibility put forward by one group that the inscriptions are merely
a set of magical or mystical symbols without any inherent meaning~\citep{Farmer04}. 
However, further analysis is needed to conclude whether the sequences
represent writing in a formal sense. This is particularly difficult as
there is no consensus about the definition of writing. As a standard
textbook on the subject mentions, ``every attempt at a single
universal definition of writing runs the risk of being either ad hoc
or anachronistic, or informed by cultural bias''~\citep{Coulmas03}.
While it can be broadly defined as a system of intercommunication by
means of conventional visible marks~\citep{Gelb63,Sampson85}, often
writing tends to be narrowly defined as a means of efficiently
encoding speech even though there is no writing system that can record
the entire linguistic structure of speech~\citep{Trigger04}.
In fact, a conception of what constitutes writing is critically
contingent upon the historical and cultural circumstances which gives
rise to the assumptions underlying such a
conception~\citep{Coulmas03}. As Coe has pointed out, the refusal to
recognize Mayan glyphs as writing because of pre-conceived notions
about what ``writing'' should be, proved to be one of the biggest
obstacles to its eventual decipherment~\citep{Coe92}.
Similarly, Ventris' decipherment of Linear B was challenged for a long
time because a writing system that leaves out endings and includes only
word stems seems strange from the point of view of modern alphabetic
writing. However, it was primarily used for ``recording accounts,
inventories and similar brief notes; there is no example of continuous
prose, \ldots the script is appropriate to its actual use, which is no
more than an elaborate kind of mnemonic device.''~\citep{Chadwick92}.

This brings us again to the point mentioned earlier in the paper that
early writing was never used for recording spoken language.
One of the objections sometimes put forward to the notion of Indus
inscriptions being a form of
writing is that there are a large number of sequences of short length.
However, many early writing systems exhibit such brevity. For example,
the written language of early Sumerian documents is very restricted
and there are no sign sequences that can be interpreted as expressions
larger than individual words. Another example is early Egyptian
writing seen in inscriptions obtained
from artifacts in royal burials dating from the late predynastic
period (c. 3200 BCE)~\citep{Baines04}.
A few hundred tags made of bone and ivory which bear around forty
different inscription types have been found in the tombs. The majority
of these tags have two hieroglyphs. Also, more than a hundred ceramic
jars have been discovered
which bear large single or paired signs painted on their outer
surfaces.
Yet another example comes from the earliest examples of Chinese
writing, viz., the very brief inscriptions on bronze ritual vessels
from the
Anyang period, belonging to the last two centuries of the 2nd
millennium BC~\citep{Bagley04}. The majority of these inscriptions
comprise only a clan sign and an ancestor dedication.
Indeed, brevity seems to be a common feature of most examples of
early
writing. This could be because the main use of writing was as a mean
of maintaining accounts, lists and other economic records. For
example, only about $15\%$ of the old Sumerian inscriptions of the
late Uruk and Jemdet Nasr periods ($3300-2900$ BCE), the period during
which the Sumerian writing system took shape, have non-economic
subject matter~\citep{Coulmas03}.

In fact, even today it is possible to see examples of such use of
writing that can result in extremely short sequences. For example, a
package may be marked by one or few signs (e.g., a numeral or an
initial consisting of alphabetic characters) in order to distinguish
it from others in terms of content, ownership, origin or destination. 
Another example is that of an inventory of a group of
commodities using a set of tally marks. Thus, very
short sequences may suggest the application of a writing system in a
specialized or restricted context, most possibly economic.
One should bear in mind that many of the shortest Indus
inscriptions contain signs comprising multiple vertical bars
resembling tally marks and which have sometimes been hypothesised to
represent numerals. Indeed, the possibly economic nature of the Indus
inscriptions have been independently suggested by evidence that many
of the seals were used to impress clay-tag sealings
that were affixed to packages~\citep{Wells10}. This suggests that the
Indus inscriptions share several characteristic features with early writing
systems rather than being an
anomaly.

\vspace{1cm}

\noindent
{\bf Acknowledgments}

We would like to thank P.~P. Divakaran, M. Vahia, N. Yadav, H.
Jogelkar and I. Mahadevan for discussions at the initial stage of the
project.


\vspace{1cm}

\appendix

\noindent
{\bf Appendix}

The Appendix comprising four sheets at the end of the paper contains a 
sign list of the W09IMSc data-set. Each sign has been numbered with a 3 digit ID
between 001 and 958. Below the ID number, the frequency of occurrence of the sign in the
W09IMSc data is indicated. 
%It may be obtained by sending an e-mail request to sitabhra$@$imsc.res.in.

\begin{figure*}[htbp]
\centering
\includegraphics[width=0.99\linewidth,clip]{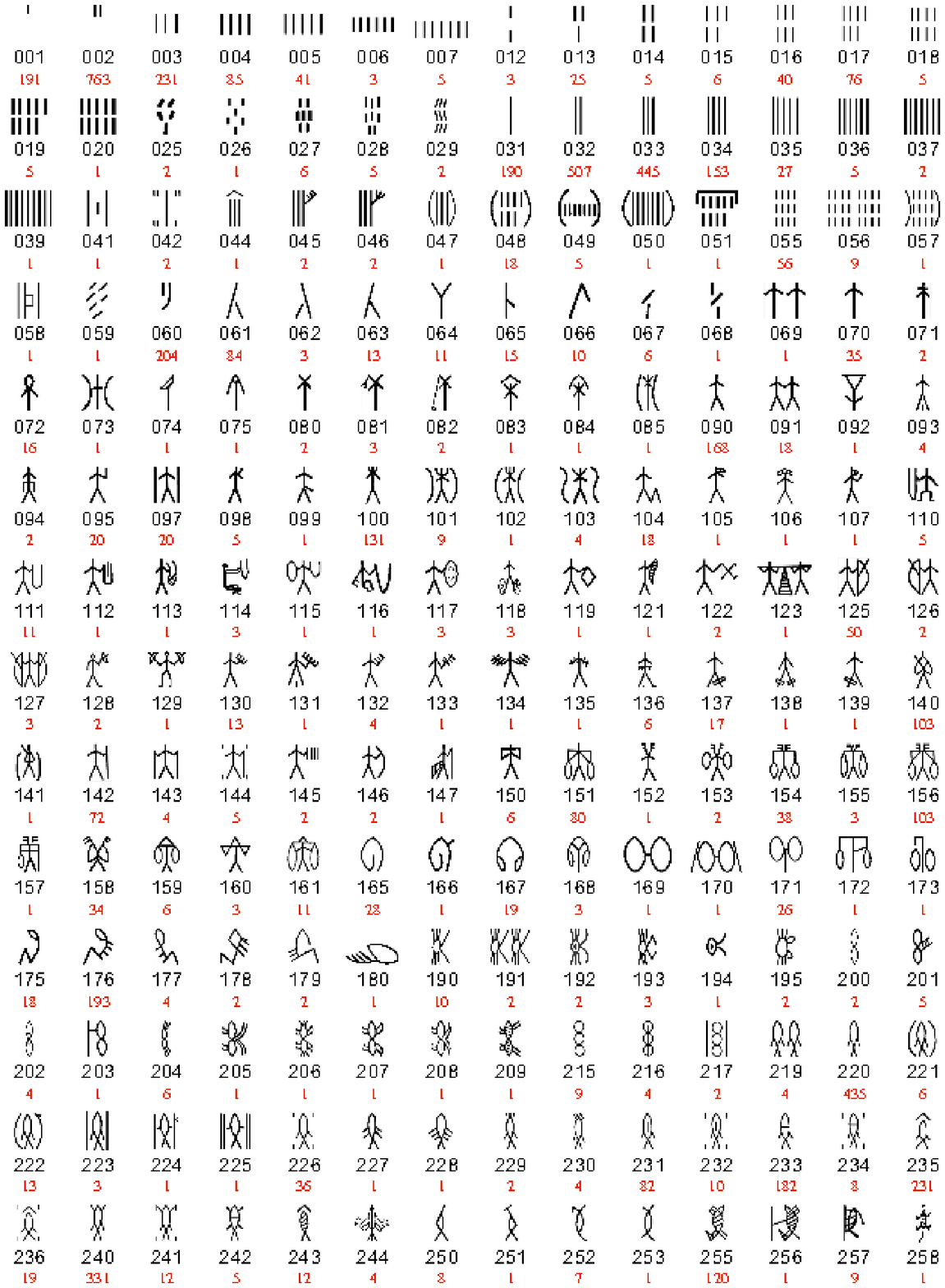}
%\caption{Appendix }
\label{Appendix_pg1}
\label{}       % Give a unique label
%\vspace{-0.15cm}
\end{figure*}

\begin{figure*}
\centering
\includegraphics[width=0.99\linewidth,clip]{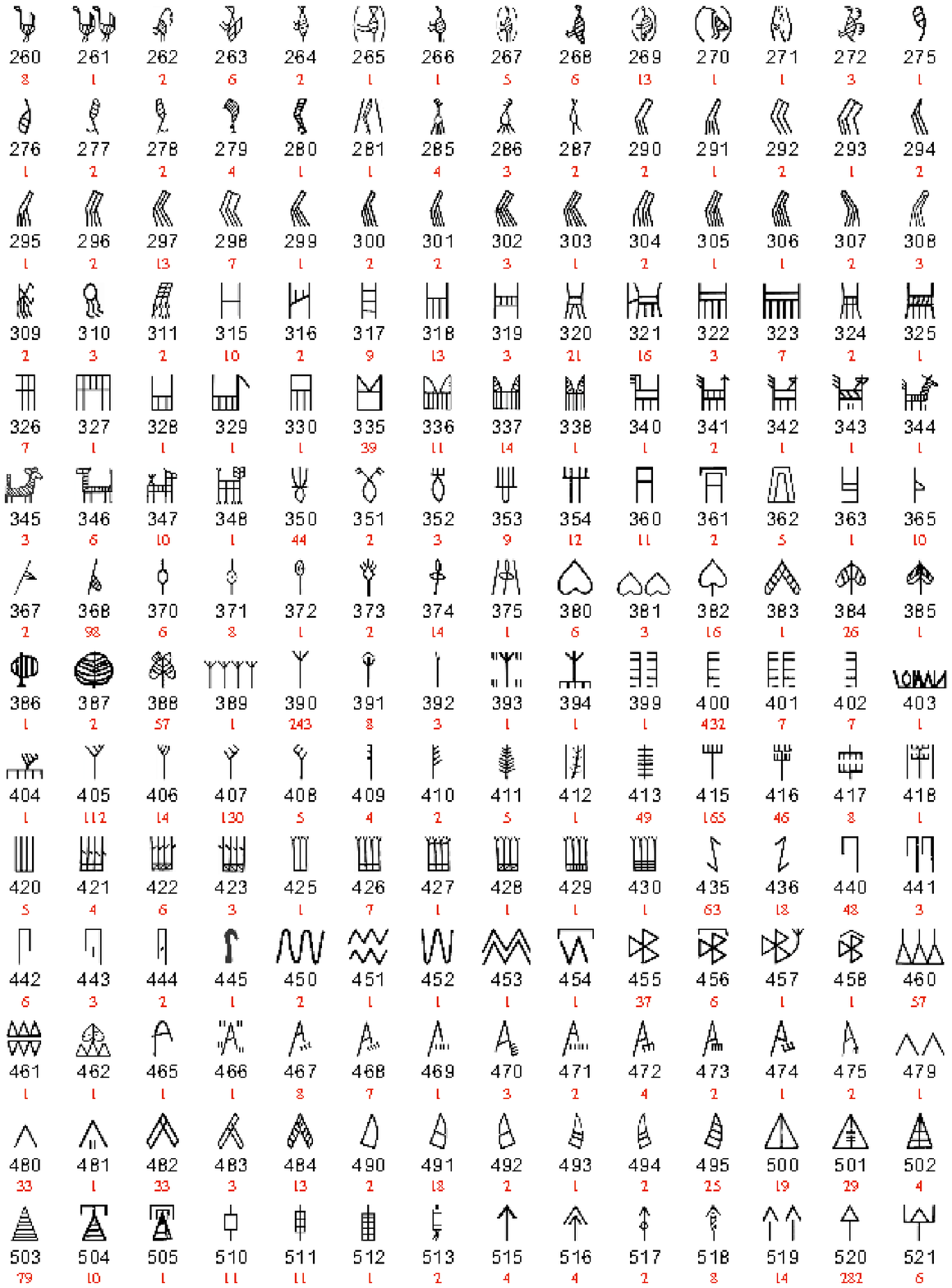}
%\caption{ }
\label{Appendix_pg2}
%\vspace{-0.15cm}
\end{figure*}

\begin{figure*}
\centering
\includegraphics[width=0.99\linewidth,clip]{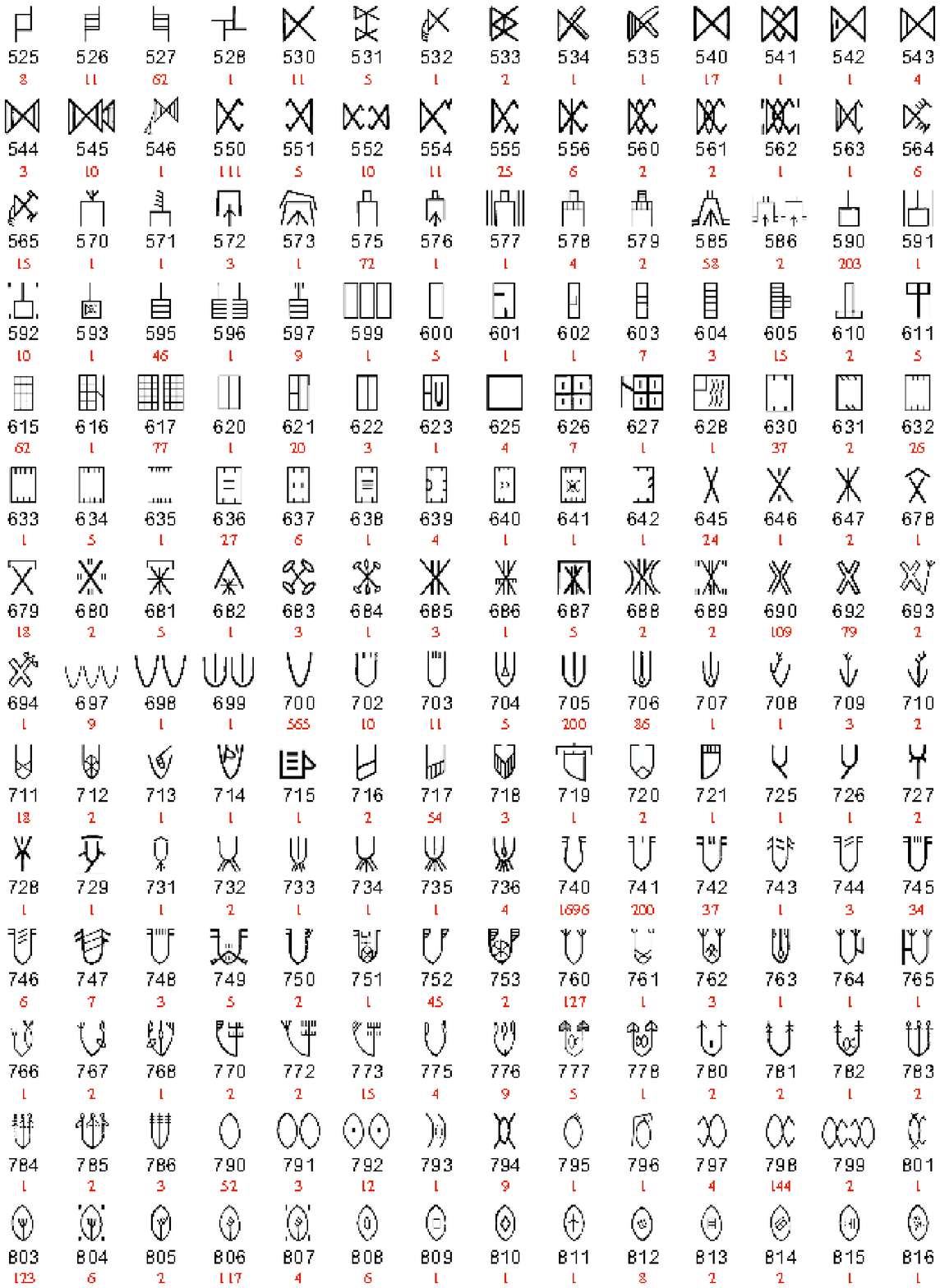}
%\caption{ }
\label{Appendix_pg3}
%\vspace{-0.15cm}
\end{figure*}

\begin{figure*}
\centering
\includegraphics[width=0.99\linewidth,clip]{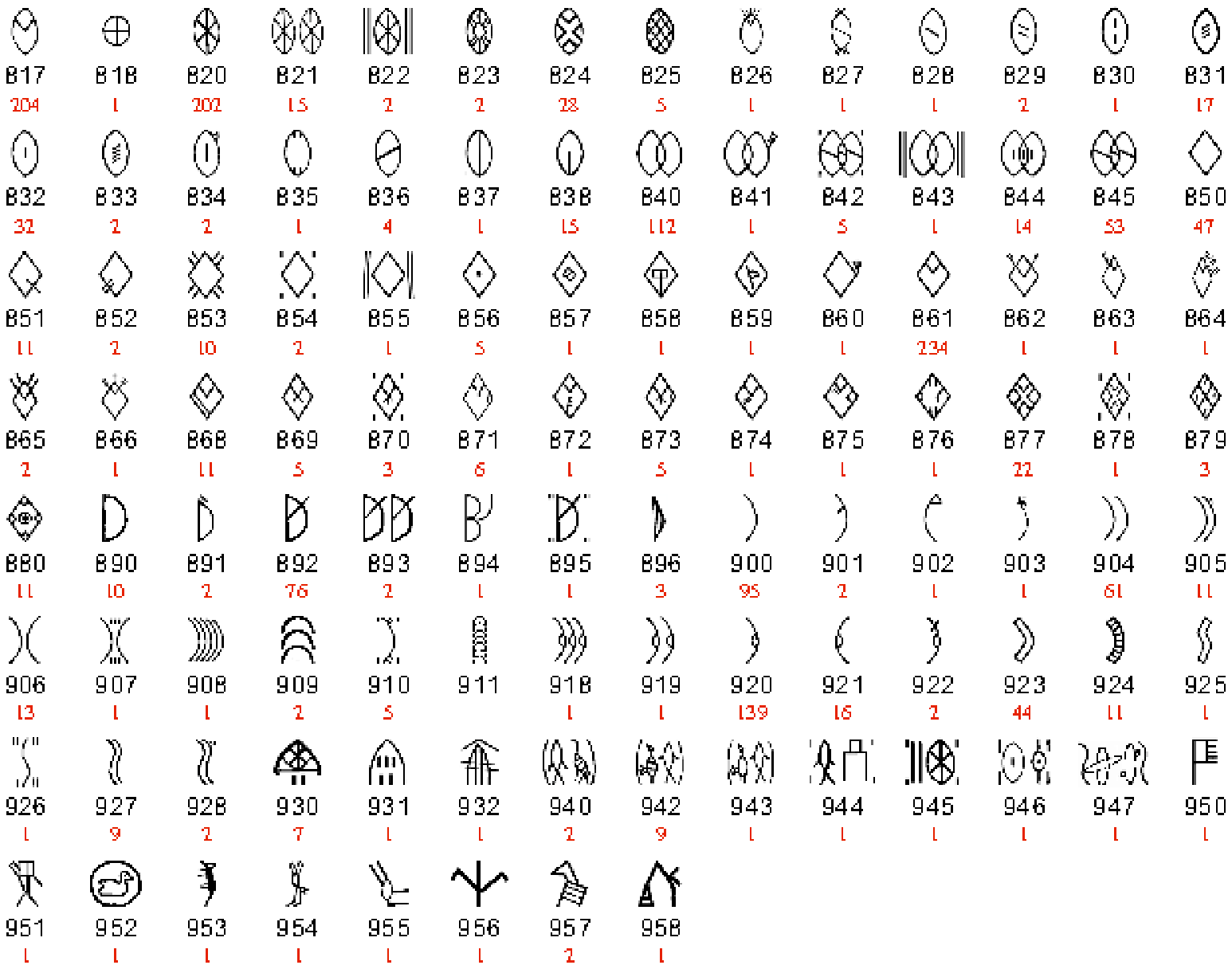}
%\caption{ }
\label{Appendix_pg4}
%\vspace{-0.15cm}
\end{figure*}


\begin{thebibliography}{00}
\bibitem{note1}
Note that the well-known Zipf
power law-frequency distribution observed in languages is defined with respect to word occurrence. Letter frequency
distributions for alphabetic scripts do not appear to fit a long-tailed distribution, partly because of the small number
of distinct signs in an alphabet. It is, at this point, premature to speculate whether the signs in the Indus inscriptions
are analogous to ``words" or ``letters".
\end{thebibliography}
\end{document}